# Is Neural Network a Reliable Forecaster on Earth? A MARS Query!

Ajith Abraham & Dan Steinberg[†]

School of Computing & Information Technology
Monash University, Churchill 3842, Australia
Email: ajith.abraham@infotech.monash.edu.au

[†] Salford Systems Inc
8880 Rio San Diego, CA 92108, USA
Email:dstein@salford-systems.com

**Abstract:** Long-term rainfall prediction is a challenging task especially in the modern world where we are facing the major environmental problem of global warming. In general, climate and rainfall are highly non-linear phenomena in nature exhibiting what is known as the "butterfly effect". While some regions of the world are noticing a systematic decrease in annual rainfall, others notice increases in flooding and severe storms. The global nature of this phenomenon is very complicated and requires sophisticated computer modeling and simulation to predict accurately. In this paper, we report a performance analysis for Multivariate Adaptive Regression Splines (MARS) [1] and artificial neural networks for one month ahead prediction of rainfall. To evaluate the prediction efficiency, we made use of 87 years of rainfall data in Kerala state, the southern part of the Indian peninsula situated at latitude-longitude pairs (8º29' N - 76º57' E). We used an artificial neural network trained using the scaled conjugate gradient algorithm. The neural network and MARS were trained with 40 years of rainfall data. For performance evaluation, network predicted outputs were compared with the actual rainfall data. Simulation results reveal that MARS is a good forecasting tool and performed better than the considered neural network.

## 1. Introduction

The parameters that are required to predict rainfall are enormously complex and subtle even for a short time period. The period over which a prediction may be made is generally termed the event horizon and in best results, this is not more than a week's time. It has been noted that the fluttering wings of a butterfly at one corner of the globe may ultimately cause a tornado at another geographically far away place. Edward Lorenz (a meteorologist at MIT) discovered this phenomenon in 1961, which is popularly known as the butterfly effect [9]. In our research, we aim to find out how well MARS and neural network models are able to capture the periodicity in these patterns so that long-term predictions can be made [7]. This would help one to anticipate the general pattern of rainfall in the coming years with some degree of confidence.

We used an artificial neural network using the scaled conjugate gradient algorithm and MARS for predicting the rainfall time series. Both models were trained on the on the rainfall data corresponding to a certain period in the past and predictions were made over some other period. In Section 2, we present some theoretical background on MARS followed by a discussion of artificial neural networks in Section3. In section 4, the experimental set up is explained followed by discussions and simulation results. Conclusions are provided at the end of the paper.

## 2. Multivariate Adaptive Regression Splines (MARS)

Splines can be considered an innovative mathematical process for complicated curve drawings and function approximation. Splines find ever-increasing application in the numerical methods, computer-aided design, and computer graphics areas. Mathematical formulae for circles, parabolas, or sine waves are easy to construct, but how does one develop a formula to trace the shape of share value fluctuations or any time series prediction problems? The answer is to break the complex shape into simpler pieces, and then use a stock formula for each piece [3]. To develop a spline the X-axis is broken into a convenient number of regions. The boundary between regions is known as a knot. With a sufficiently large number of knots virtually any shape can be well approximated. While it is easy to draw a spline in two dimensions by keying on knot locations (approximating using linear, quadratic or cubic polynomial regression etc.), manipulating the mathematics in higher dimensions is best accomplished using basis functions.

The MARS model is a spline regression model that uses a specific class of basis functions as predictors in place of the original data [1]. The MARS basis function transform makes it possible to selectively blank out certain regions of a variable by making them zero, allowing MARS to focus on specific sub-regions of the data. MARS excels at finding optimal variable transformations and interactions, as well as the complex data structure that often hides in high-dimensional data [2] [6].

Given the number of predictors in most data mining applications, it is infeasible to approximate a function $y=f(x)$ in a generalization of splines by summarizing $y$ in each distinct region of $x$. Even if we could assume that each predictor $x$ had only two distinct regions, a database with just 35 predictors would contain $2^{35}$ or more than 34 billion regions. This is known as the curse of dimensionality. For some variables, two regions may not be enough to track the specifics of the function. If the relationship of $y$ to some $x$'s is different in three or four regions, for example, with only 35 variables the number of regions requiring examination would be even larger than 34 billion. Given that neither the number of regions nor the knot locations can be specified a priori, a procedure is needed that accomplishes the following:

- judicious selection of which regions to look at and their boundaries, and
- judicious determination of how many intervals are needed for each variable.

A successful method of region selection will need to be adaptive to the characteristics of the data. Such a solution will probably reject quite a few variables (accomplishing variable selection) and will take into account only a few variables at a time (also reducing the number of regions). Even if the method selects 30 variables for the model, it will not look at all 30 simultaneously. Similar simplification is accomplished by a decision tree (e.g., at a single node, only ancestor splits are being considered; thus, at a depth of six levels in the tree, only six variables are being used to define the node).

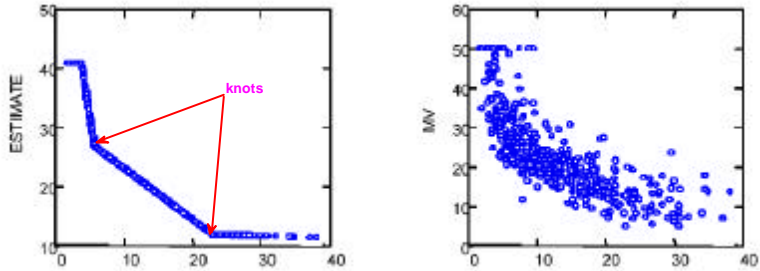

**Figure 1.** MARS data estimation using spines and knots (actual data on the right)

### MARS Smoothing, Splines, Knots Selection and Basis Functions

A key concept underlying the spline is the knot, which marks the end of one region of data and the beginning of another. Thus, the knot is where the behavior of the function changes. Between knots, the model could be global (e.g., linear regression). In a classical spline, the knots are predetermined and evenly spaced, whereas in MARS, the knots are determined by a search procedure. Only as many knots as needed are included in a MARS model. If a straight line is a good fit, there will be no interior knots. In MARS, however, there is always at least one "pseudo" knot that corresponds to the smallest observed value of the predictor. Figure 1 depicts a MARS spline with three knots.

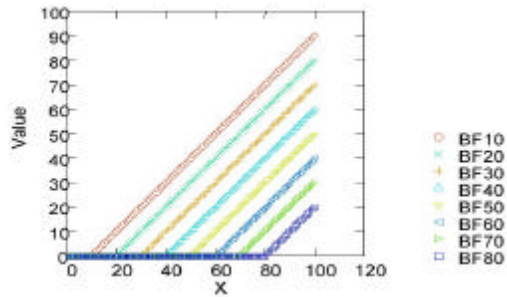

**Figure 2.** Variations of basis functions for c = 10 to 80

Finding the one best knot in a simple regression is a straightforward search problem: simply examine a large number of potential knots and choose the one with the best $R^2$. However, finding the best pair of knots requires far more computation, and finding the best set of knots when the actual number needed is unknown is an even more challenging task. MARS finds the location and number of needed knots in a forward/backward stepwise fashion. First a model which is clearly overfit with too many knots is generated, then, those knots that contribute least to the overall fit are removed. Thus, the forward knot selection will include many incorrect knot locations, but these erroneous knots should eventually, be deleted from the model in the backwards pruning step (although this is not guaranteed).

Thinking in terms of knot selection works very well to illustrate splines in one dimension; however, this context is unwieldy for working with a large number of variables simultaneously. Both concise notation and easy to manipulate programming expressions are required. It is also not clear how to construct or represent interactions using knot locations. In MARS, Basis Functions (BFs) are the machinery used for generalizing the search for knots. BFs are a set of functions used to represent the information contained in one or more variables. Much like principal components, BFs essentially re-express the relationship of the predictor variables with the target variable. The hockey stick BF, the core building block of the MARS model is often applied to a single variable multiple times. The hockey stick function maps variable $X$ to new variable $X^*$:

$\max(0, X-c)$, or
$\max(0, c - X)$.

In the first form, $X^*$ is set to 0 for all values of $X$ up to some threshold value c and $X^*$ is equal to $X$ for all values of $X$ greater than c. (Actually $X^*$ is equal to the amount by which $X$ exceeds threshold c.) The second form generates a mirror image of the first. Figure 2 illustrates the variation in BFs for changes of c values (in steps of 10) for predictor variable $X$, ranging from 0 to 100. MARS generates basis functions by searching in a stepwise manner. It starts with a constant in the model and then begins the search for a variable-knot combination that improves the model the most (or, alternatively, worsens the model the least). The improvement is measured in part by the change in Mean Squared Error (MSE). Adding a basis function always reduces the MSE. MARS searches for a pair of hockey stick basis functions, the primary and mirror image, even though only one might be linearly independent of the other terms. This search is then repeated, with MARS searching for the best variable to add given the basis functions already in the model. The brute search process theoretically continues until every possible basis function has been added to the model.

In practice, the user specifies an upper limit for the number of knots to be generated in the forward stage. The limit should be large enough to ensure that the true model can be captured. A good rule of thumb for determining the minimum number is three to four times the number of basis functions in the optimal model. This limit may have to be set by trial and error.

## 3. Artificial Neural Network (ANN)

ANN is an information-processing paradigm inspired by the way the densely interconnected, parallel structure of the mammalian brain processes information. Learning in biological systems involves adjustments to the synaptic connections that exist between the neurons [5]. Learning typically occurs by example through training, where the training algorithm iteratively adjusts the connection weights (synapses).

These connection weights store the knowledge necessary to solve specific problems. Backpropagation (BP) is one of the most famous training algorithms for multilayer perceptrons. BP is a gradient descent technique to minimize the error *E* for a particular training pattern. For adjusting the weight ($w_{ij}$) from the *i*-th input unit to the *j*-th output, in the batched mode variant the descent is based on the gradient $\nabla E$ ($\frac{dE}{dw_{ij}}$) for the total training set:

$$\Delta w_{ij}(n) = -e* \frac{dE}{dw_{ij}} + a* \Delta w_{ij}(n-1) \tag{1}$$

The gradient gives the direction of error *E*. The parameters ε and α are the learning rate and momentum respectively [4].

In the Conjugate Gradient Algorithm (CGA) a search is performed along conjugate directions, which produces generally faster convergence than steepest descent directions. A search is made along the conjugate gradient direction to determine the step size, which will minimize the performance function along that line. A line search is performed to determine the optimal distance to move along the current search direction. Then the next search direction is determined so that it is conjugate to previous search direction. The general procedure for determining the new search direction is to combine the new steepest descent direction with the previous search direction. An important feature of the CGA is that the minimization performed in one step is not partially undone by the next, as it is the case with gradient descent methods. The key steps of the CGA is summarized as follows:

- Choose an initial weight vector $w_i$.
- Evaluate the gradient vector $g_1$, and set the initial search direction $d_1 = -g_1$
- At step j, minimize $E(w_j + ad_j)$ with respect to *a* to give $w_{j+1} = w_j + a_{min}d_j$)
- Test to see if the stopping criterion is satisfied.
- Evaluate the new gradient vector $g_{j+1}$
- Evaluate the new search direction using $d_{j+1} = -g_{j+1} + \beta_j d_j$. The various versions of conjugate gradient are distinguished by the manner in which the constant $\beta_j$ is computed.

An important drawback of CGA is the requirement of a line search, which is computationally expensive. The Scaled Conjugate Gradient Algorithm (SCGA) is basically designed to avoid the time-consuming line search at each iteration. SCGA combine the model-trust region approach, which is used in the Levenberg-Marquardt algorithm with the CGA. Detailed step-by-step descriptions of the algorithm can be found in Moller[8].

## 4. Experimental Setup Using Neural Networks and MARS

The 87 years (1893-1980) rainfall data was standardized and the first 40 years was used for training the prediction models and the remaining for testing purposes. We used 12 inputs (previous 4 years rainfall data) to predict the amount of rain to be expected in each month of the fifth year. Experiments were carried out on a Pentium II 450MHz machine and the codes were executed using MATLAB. Test data was presented to the network and the output from the network was compared with the actual data in the time series.

- **ANN – SCG algorithm**

    We used a feedforward neural network with two hidden layers consisting of 12 neurons each. We used log-sigmoidal activation function for the hidden neurons. The training was terminated after 600 epochs.

- **MARS**

    We increased the number of basis functions in steps of five and selected 1 as the setting of minimum observation between knots. To obtain the best possible prediction results (lowest RMSE), we sacrificed speed (minimum completion time).

- **Performance and results achieved**

    Figure 3 illustrates the training performance of the proposed neural network. Table 2 summarizes the comparative performances of MARS and neural network. Figure 4 shows the change in RMSE values for change in number of basis functions.

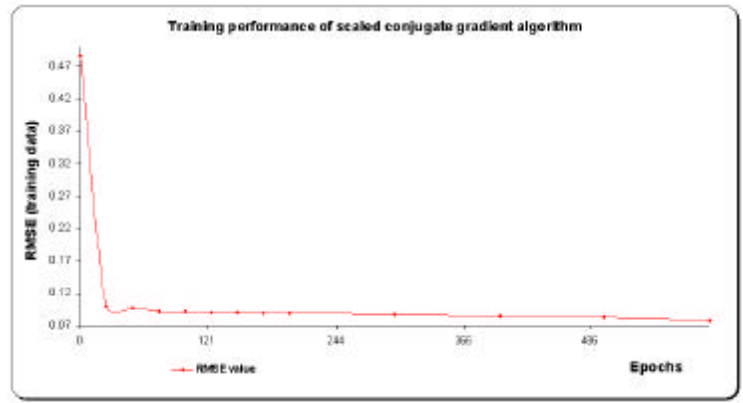

**Figure 3**. Training performance of SCGA for 600 epochs

**Table1:** Training and testing performance comparison

| Model | RMSE | | Epochs | Training time (seconds) |
|---|---|---|---|---|
| | **Training set** | **Test set** | | |
| **MARS** | 0.0990 | 0.0832 | - | 5 |
| **ANN-SCG** | 0.0780 | 0.0923 | 600 | 90 |

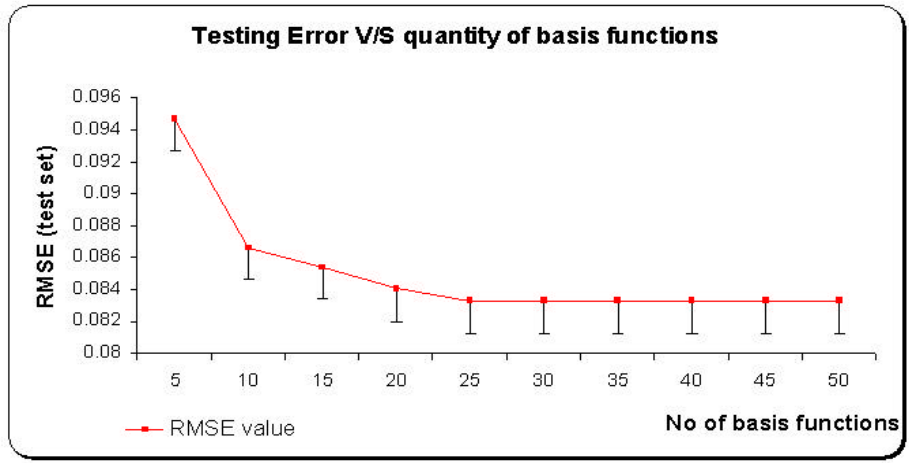

**Figure 4**. MARS: Test set RMSE convergence

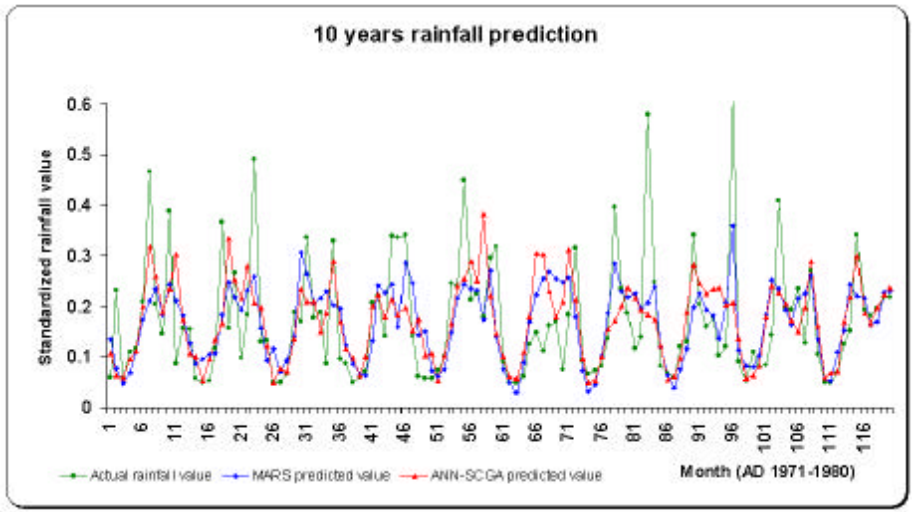

**Figure 5.** Test results showing one-month ahead prediction of rainfall for 10 years using MARS and ANN-SCGA.

## 5. Conclusion

In this paper, we attempted to forecast rainfall (one month ahead) using MARS and a neural network trained using SCGA. As the RMSE values on test data are comparatively less, the prediction models are reliable. As evident from the entire test results (for 47 years), there have been few deviations of the predicted rainfall value from the actual. In some cases it is due to delay in the actual commencement of monsoon, EI-Nino Southern Oscillations (ENSO) resulting from the pressure oscillations between the tropical Indian Ocean and the tropical Pacific Ocean and their quasi periodic oscillations [10]

Prediction results reveal that MARS is capable of outperforming neural networks in terms of performance time and lowest RMSE. In our experiments, we used only 40 years training data to evaluate the learning capability of the models. Network performance could have been further improved by providing more training data. It will be interesting to study further the robustness of MARS when compared to neural networks.

## Acknowledgement

The authors wish to express their sincere thanks to Professor K Mohankumar of the Department of Atmospheric Sciences of the Cochin University of Science and Technology for providing the rainfall database.